\documentclass{ieeetj}
\usepackage{cite}
\usepackage{amsmath,amssymb,amsfonts}
\usepackage{algorithmic}
\usepackage{graphicx,color}
\usepackage{textcomp}
\usepackage{xcolor}
\usepackage{hyperref}
\hypersetup{hidelinks=true}
\usepackage{algorithm,algorithmic}
\def\BibTeX{{\rm B\kern-.05em{\sc i\kern-.025em b}\kern-.08em
    T\kern-.1667em\lower.7ex\hbox{E}\kern-.125emX}}
\AtBeginDocument{\definecolor{tmlcncolor}{cmyk}{0.93,0.59,0.15,0.02}\definecolor{NavyBlue}{RGB}{0,86,125}}

\usepackage{cite}
\usepackage{amsmath,amssymb,amsfonts}
\usepackage{algorithmic}
\usepackage{graphicx}
\usepackage{textcomp}
\usepackage[x11names]{xcolor}

\usepackage{array}
\usepackage{mathtools}
\usepackage{tcolorbox}
\usepackage{color}
\usepackage{theorem}
\usepackage{amssymb}
\usepackage{caption}
\usepackage{subcaption}
\usepackage{cite,hyperref}
\usepackage{cases}
\usepackage{url}
\usepackage{algorithmic}
\usepackage{algorithm}
\usepackage{multirow}
\usepackage{hhline}
\input{my_styles.sty}
\usepackage{symbolDef}
\usepackage{booktabs}
\usepackage{subcaption}
\usepackage{pgfplots}

\usepackage{tikz}
\usetikzlibrary{shapes,arrows}
\usepackage{moresize}
\usepackage{pgfplots}
\usepackage{wrapfig}
\pgfplotsset{compat=1.17}
\pgfplotstableset{col sep=comma}

\usepackage{xcolor}
\definecolor{darkblue}{RGB}{0,0,139}

\usetikzlibrary{matrix,positioning,calc,decorations.pathreplacing}

\providecommand{\Her}{\mathsf{H}}                 

\usepackage{theorem}
\newtheorem{theorem}{Theorem}
\newtheorem{assumption}{Assumption}
\newtheorem{lemma}{Lemma}

\def\OJlogo{\vspace{-4pt}$<$Society logo(s) and publication title will appear here.$>$}
\def\seclogo{\vspace{10pt}$<$Society logo(s) and publication title will appear here.$>$}

\begin{document}
\receiveddate{XX Month, XXXX}
\reviseddate{XX Month, XXXX}
\accepteddate{XX Month, XXXX}
\publisheddate{XX Month, XXXX}
\currentdate{XX Month, XXXX}
\doiinfo{XXXX.2022.1234567}

\markboth{}{Cavallo {et al.}}

\title{Spatiotemporal Kronecker Covariance Neural Networks}

\author{Andrea Cavallo, Student Member, IEEE, Athanasios Georgoutsos,\\ and Elvin Isufi, Senior Member, IEEE}
\affil{Delft University of Technology, Delft, Netherlands}
\corresp{Corresponding author: Andrea Cavallo (email: a.cavallo@tudelft.nl).}
\authornote{This work was supported in part by the TU Delft AI Labs programme, NWO OTP GraSPA proposal \#19497, NWO VENI proposal 222.032, and by the SURE-AI Centre grant \#357482, Research Council of Norway.
This work is not related to Athanasios Georgoutsos' position at Amazon.com Inc.
}

\begin{abstract}
Multivariate time series contain complex patterns that span across both space and time. While covariance-based statistical tools like spatiotemporal Principal Component Analysis (ST-PCA) help identify these patterns, they are limited to linear operations and prone to estimation errors with limited data. Recent covariance-based spatiotemporal neural networks offer more stable, non-linear alternatives, but they ignore correlations across different time steps. To solve this, we introduce the  Kronecker coVariance Neural Network (KVNN), a temporal graph neural network that represents the spatiotemporal covariance matrix via a sum of Kronecker products where spatial and temporal dependencies are decoupled. By implementing filtering operations on spatial and temporal components, KVNNs achieve expressive processing capabilities, admit a rigorous spectral analysis, and are provably stable to finite-sample estimation errors, ultimately addressing all of ST-PCA's limitations.
We show on five real-world datasets that KVNNs achieve strong forecasting performance, often requiring significantly fewer trainable parameters than competitive methods, and are consistent under estimation noise.
\end{abstract}

\begin{IEEEkeywords}
Covariance Neural Networks, Spatiotemporal Principal Component Analysis
\end{IEEEkeywords}


\maketitle

\section{INTRODUCTION}

Multivariate time series exhibit intricate spatiotemporal correlations whose successful estimation and processing is key for forecasting tasks as well as for modeling and understanding spatiotemporal behaviors~\cite{Lin_2018,Lei_2019, Karevan_2020, Hewage_2020, Ismaeel_2023}. 
ST-PCA offers a prominent tool to identify spatiotemporal patterns, enhance understanding and tractability of multivariate time series~\cite{Mao_2019, Krzysko_2024}. Usually, ST-PCA relies on the eigendecomposition of a large spatiotemporal covariance matrix, either built directly from an extended space-time covariance matrix~\cite{Jombart_2008, Krzysko_2024}, or on decoupled parametrizations across space and time. It has found applications in traffic flow forecasting~\cite{Mao_2019}, galaxy dynamics~\cite{Weinberg_2020}, and online change-point detection~\cite{Alanqary_2021}, among others.
Despite its informativeness and widespread use, ST-PCA suffers from two crucial limitations. First, it has limited expressivity, since it only offers linear processing of covariance information, and is task-agnostic. Second, it is sensitive to finite-sample errors, which prevents stable performance across repeated trials with different sampling noise.

A covariance-based alternative to enhance expressivity and stability are spatiotemporal coVariance Neural Networks (STVNNs)~\cite{cavallo2024stvnn}; temporal graph neural networks operating on sample covariance as a proxy of a graph structure of the data. These are based on the VNN principle~\cite{sihag2022covariance,sihag2026learning}, where the covariance matrix of the data is used as an inductive bias in a graph neural network trained end-to-end to solve a downstream task, and extend VNNs to include dynamic information via a temporal window of past data. In this way, STVNNs are highly expressive via non-linear covariance processing and are provably more stable to finite-sample estimation errors than PCA. However, they only consider covariances in individual snapshots, thus ignoring temporally delayed correlations. This neglects crucial spatiotemporal information and weakens the connection with ST-PCA.

To address these limitations, we introduce the Kronecker coVariance Neural Network (KVNN), a temporal graph neural network that operates on the spatiotemporal covariance matrix as a graph shift operator, thus accounting for correlations across space and time, and decomposes such covariance as a sum of Kronecker terms, representing spatial and temporal connections separately. 
This covariance representation allows decoupled processing of different temporal delays, overcoming STVNN's limitations.
To further improve efficiency and interpretability, we integrate a group-sparsity penalty during training, enabling automatic pruning of uninformative spatiotemporal blocks for the task at hand.

We propose two practical implementations of KVNNs. The first one relies on fixed temporal matrices and estimated covariances at fixed delays, which admits a rigorous analysis of their spectral behavior and stability to finite-sample errors.
The second one, instead, uses a low-rank Kronecker covariance estimate to model long-term dependencies with limited Kronecker terms, thus increasing parameter efficiency.

\smallskip \noindent \textbf{Related works.}
The proposed approach could be seen as a deep learning solution to process multivariate time series using the spatiotemporal covariance as an inductive bias. Alternative deep learning-based methods for this task also exist, including recurrent models~\cite{che2018recurrent,lai2018modeling}, graph-time neural networks~\cite{Li_2017, Wu_2019, Wu_2020, sabbaqi2023graph, jin2024survey} and transformers~\cite{Zhou_2021,zhang2023crossformer,zerveas2021transformer}.
However, these methods either ignore the spatial structure, assume a known graph structure, or they learn it via free parameters or generic attention scores rather than employing an explicit statistical prior, missing a connection to spatiotemporal PCA and its interpretability and stability guarantees.

The VNN principle that our method relies on was introduced in~\cite{sihag2022covariance,sihag2026learning} and has received increasing attention in methodological extensions~\cite{cavallo2024sparsecovarianceneuralnetworks,cavallo2026covariance,roy2026covariance} and applications~\cite{sihag2024explainable,sihag2024transferability,sihag2025disentangling}. STVNNs~\cite{cavallo2024stvnn} apply this principle to temporal data, but ignore delayed correlations. 
We proposed a preliminary version of KVNNs, termed lagged VNNs (LVNNs), in~\cite{Georgoutsos2025}. These, however, lack expressivity to separately process delayed correlations, and are computationally inefficient as they do not consider Kronecker covariance representations.

\smallskip \noindent \textbf{Contributions.}
Our contributions are threefold:

\noindent \textbf{(C1)}
We propose the KVNN, a neural network that operates on the Kronecker representation of the spatiotemporal covariance matrix, processing lagged covariances in an expressive and efficient manner.

\noindent \textbf{(C2)}
We provide a spectral analysis of KVNN, and we prove that its asymptotic stability against finite-sample estimation errors improves over ST-PCA.

\noindent \textbf{(C3)}
We provide extensive numerical experiments on five real-world datasets, demonstrating that KVNNs achieve highly competitive forecasting performance while benefiting from the parameter efficiency of group-sparse and low-rank representations.


\begin{figure*}[htbp]
\centering
\begin{subfigure}[b]{0.65\textwidth}
\centering
\resizebox{\linewidth}{!}{%
\begin{tikzpicture}[
    ampersand replacement=\&,
    matgrid/.style={
        matrix of math nodes,
        nodes={draw, minimum size=0.5cm, anchor=center, inner sep=1pt, font=\footnotesize},
        column sep=-\pgflinewidth, row sep=-\pgflinewidth, draw,
    },
    label node/.style={font=\small, align=center},
    dim label/.style={font=\footnotesize, color=gray},
    op/.style={font=\Large},
    diag/.style={fill=blue!12},
    offd/.style={fill=blue!4},
]
\matrix [matgrid] (C_mat)
{
|[diag]| \mathbf{C}_{0,0}   \& |[offd]| \mathbf{C}_{0,1}   \& |[offd]| \dots  \& |[offd]| \mathbf{C}_{0,T-1} \\
|[offd]| \mathbf{C}_{1,0}   \& |[diag]| \mathbf{C}_{1,1}   \& |[offd]| \dots  \& |[offd]| \mathbf{C}_{1,T-1} \\
|[offd]| \vdots             \& |[offd]| \vdots             \& |[diag]| \ddots \& |[offd]| \vdots \\
|[offd]| \mathbf{C}_{T-1,0} \& |[offd]| \mathbf{C}_{T-1,1} \& |[offd]| \dots  \& |[diag]| \mathbf{C}_{T-1,T-1} \\
};
\node[label node, above=0.1cm of C_mat] {$\tilde{\mathbf{C}} \in \mathbb{R}^{NT \times NT}$};
\draw[decorate,decoration={brace,amplitude=3pt,mirror}]
    ($(C_mat.south west)+(0,-0.06)$) -- ($(C_mat.south east)+(0,-0.06)$)
    node[midway,below=2pt,dim label] {$T$ blocks};

\node[op] (equals)  [right=0.1cm of C_mat]  {$=$};
\node[op] (sum_sym) [right=0.05cm of equals] {$\displaystyle\sum_{r=1}^R$};

\matrix [matgrid] (Ar_mat) [right=0.1cm of sum_sym]
{
|[fill=red!15]| a_{00} \& |[fill=red!5]|  a_{01} \& \dots  \\
|[fill=red!5]|  a_{10} \& |[fill=red!15]| a_{11} \& \dots  \\
\vdots                 \& \vdots                 \& \ddots \\
};
\node[label node, above=0.1cm of Ar_mat] {$\mathbf{A}_r$ (temporal)};
\draw[decorate,decoration={brace,amplitude=3pt,mirror}]
    ($(Ar_mat.south west)+(0,-0.06)$) -- ($(Ar_mat.south east)+(0,-0.06)$)
    node[midway,below=2pt,dim label] {$T \times T$};

\node[op] (kron_sym) [right=0.1cm of Ar_mat] {$\otimes$};
\node[op] (Br)       [right=0.1cm of kron_sym] {$\mathbf{B}_r$};
\node[label node, above=0.1cm of Br] {(spatial)};
\node[dim label,  below=0.1cm of Br] {$N \times N$};

\node[op] (arrow) [right=0.1cm of Br] {$\to$};

\matrix [matgrid] (Kron_expl) [right=0.1cm of arrow]
{
|[fill=red!15]| a_{00}\mathbf{B}_r \& |[fill=red!5]|  a_{01}\mathbf{B}_r \& \dots  \\
|[fill=red!5]|  a_{10}\mathbf{B}_r \& |[fill=red!15]| a_{11}\mathbf{B}_r \& \dots  \\
\vdots                             \& \vdots                             \& \ddots \\
};
\node[label node, above=0.1cm of Kron_expl] {block structure};
\draw[decorate,decoration={brace,amplitude=3pt,mirror}]
    ($(Kron_expl.south west)+(0,-0.06)$) -- ($(Kron_expl.south east)+(0,-0.06)$)
    node[midway,below=2pt,dim label] {$T$ blocks};
\end{tikzpicture}%
}
\label{fig:kronecker_structure}
\end{subfigure}\hfill
{\color{gray!60}\vrule width 0.5pt}%
\hfill
\begin{subfigure}[b]{0.3\textwidth}\centering
\resizebox{\linewidth}{!}{%
\begin{tikzpicture}[
    series/.style={gray!55, thin},
    snap/.style={gray!70, dashed, thin},
    gnode/.style={circle, draw=blue!60!black, fill=blue!15, inner sep=0pt, minimum size=5pt},
    spatial/.style={green!55!black, thick},
    temporal/.style={red!70!black, thick},
    temporal long/.style={temporal, dash pattern=on 3pt off 1.5pt},
    feat/.style={font=\scriptsize, inner sep=1pt},
    cov/.style={font=\scriptsize, inner sep=1pt},
]
\foreach \x/\lab in {1.6/{t-2}, 3.4/{t-1}, 5.2/{t}} {
    \draw[snap] (\x,-0.5) -- (\x,2.4);
    \node[font=\small, below] at (\x,-0.5) {$\lab$};
}

\draw[series] plot[smooth,tension=0.8] coordinates
  {(0.9,1.45) (1.6,1.7) (2.5,1.8) (3.4,1.5) (4.3,1.7) (5.2,1.5) (5.9,1.72)};
\draw[series] plot[smooth,tension=0.8] coordinates
  {(0.9,0.95) (1.6,0.8) (2.5,0.65) (3.4,0.92) (4.3,0.75) (5.2,0.9) (5.9,0.7)};
\draw[series] plot[smooth,tension=0.8] coordinates
  {(0.9,0.15) (1.6,0.0) (2.5,0.18) (3.4,0.08) (4.3,-0.15) (5.2,0.15) (5.9,0.0)};

\foreach \n/\x/\y/\tt/\i/\side in {
    n1a/1.6/1.7/{t-2}/1/{above left},  n2a/1.6/0.8/{t-2}/2/{above left},   n3a/1.6/0.0/{t-2}/3/{above left},
    n1b/3.4/1.5/{t-1}/1/{above right}, n2b/3.4/0.92/{t-1}/2/{above right}, n3b/3.4/0.08/{t-1}/3/{above right},
    n1c/5.2/1.5/{t}/1/{right},         n2c/5.2/0.9/{t}/2/{right},          n3c/5.2/0.15/{t}/3/{right}}
    \node[gnode, label={[feat]\side:{$[\mathbf{x}_{\tt}]_{\i}$}}] (\n) at (\x,\y) {};

\foreach \c in {a,b,c} {
    \draw[spatial] (n1\c) -- (n2\c);
    \draw[spatial] (n2\c) -- (n3\c);
    \draw[spatial] (n1\c) to[bend right=35] (n3\c);
}
\foreach \p/\q in {a/b, b/c} {
    \foreach \i in {1,...,3} \draw[temporal] (n\i\p) -- (n\i\q);
    \draw[temporal] (n1\p) -- (n2\q);
    \draw[temporal] (n2\p) -- (n3\q);
    \draw[temporal] (n3\p) -- (n2\q);
}
\draw[temporal long] (n1a) to[bend left=25]  (n1c);
\draw[temporal long] (n3a) to[bend right=25] (n3c);

\draw[spatial]  (n2c) -- (n3c)
    node[cov, midway, right=2pt, text=green!55!black] {$[\mathbf{C}_{0}]_{23}$};
\draw[temporal] (n2a) -- (n2b)
    node[cov, midway, above, text=red!70!black] {$[\mathbf{C}_{1}]_{22}$};
\draw[temporal] (n1b) -- (n2c)
    node[cov, midway, sloped, above, text=red!70!black] {$[\mathbf{C}_{1}]_{12}$};
\draw[temporal long] (n1a) to[bend left=25]
    node[cov, midway, above, text=red!70!black] {$[\mathbf{C}_{2}]_{11}$} (n1c);
\end{tikzpicture}%
}
\label{fig:st_graph}
\end{subfigure}
\caption{(Left) Kronecker representation of the spatiotemporal covariance $\tilde{\mathbf{C}}$ as a sum of $R$ decoupled interactions. The temporal covariance $\mathbf{A}_r$ scales the spatial covariance $\mathbf{B}_r$ to form the block structure. (Right) Spatiotemporal covariance graph on a multivariate time series. Node $i$ at time $t$ carries the feature $[\mathbf{x}_t]_i$, edges connect nodes within a snapshot (green), and across snapshots (red). Edge weights are the covariances $[\mathbf{C}_\tau]_{ij}$ at lag $\tau$.}
\label{fig:overview}
\end{figure*}
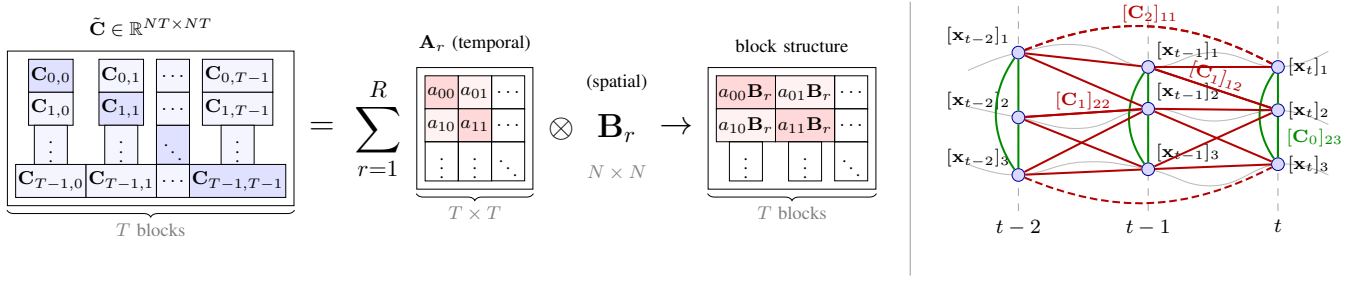

\section{SPATIOTEMPORAL COVARIANCES}

Consider a data matrix $\mtX \in \mathbb{R}^{N \times M'}$ where each column $\vcx_t \in \mathbb{R}^N$ is an observation of a zero-mean multivariate time series and the observations are in temporal order. To model the joint spatiotemporal correlations in $\mtX$, we consider a temporal window of size $T$ and we stack $T$ consecutive temporal samples into a unique vector $\vctx_t = [\vcx_{t-T+1}^\Tr, \vcx_{t-T+2}^\Tr, \dots, \vcx_{t}^\Tr]^\Tr$. 
By rearranging the data matrix $\mtX$ in this way, we get the matrix $\mttX \in \mathbb{R}^{NT \times M}$ with $M = M'/T$, which collects observations of the variable $\vctx \in \mathbb{R}^{NT}$ with covariance matrix
\begin{equation}\label{eq:cov_def}
    \mttC_0 = \mathbb{E}[\vctx \vctx^\Tr] \in \mathbb{R}^{NT\times NT}.
\end{equation}
$\mttC_0$ encodes correlations across space and time via a block structure containing $T \times T$ 
blocks $\mtC_{t_1,t_2} = \mathbb{E}[\vcx_{t_1}\vcx_{t_2}^\Tr] \in \mathbb{R}^{N \times N}$ for $t_1, t_2 = 0, \dots T-1$, i.e., the covariance among samples at time points $t_1, t_2$ (cf. Figure~\ref{fig:overview}, left).

$\mttC_0$ can be directly estimated from the samples by stacking $T$ consecutive observations and using the sample estimator
\begin{align}\label{eq:sample_cov}
    \mttC_s = \frac{1}{M} \sum_{t=1}^M \vctx_t\vctx_t^\Tr.
\end{align}
However, $\mttC_s$ is high-dimensional, thus requiring the estimation of a large number of parameters, which may lead to poor results if insufficient data is available, and causes data processing techniques relying on this matrix (e.g., spatiotemporal PCA) to be computationally inefficient. Moreover, $\mttC_s$ jointly models spatial and temporal correlations, making it difficult to interpret the resulting principal components.

Considering its spatiotemporal structure, the matrix $\mttC_0$ can be approximated as a sum of $R$ Kronecker products
\begin{equation}\label{eq:kron_cov}
    \mttC = \sum_{r=1}^R \mtA_r \otimes \mtB_r
\end{equation}
where matrices $\mtA_r \in \mathbb{R}^{T\times T}$ encode temporal relations and matrices $\mtB_r\in \mathbb{R}^{N\times N}$ spatial correlations (cf. Figure~\ref{fig:overview}, left).
This representation decouples the spatial and the temporal contributions, allowing for better interpretability and various estimators based on the data assumptions.
In general, $\mttC \neq \mttC_0$, but for specific choices of Kronecker terms, equality holds.
We present different choices of $\mtA_r, \mtB_r$ below.

\subsection{WIDE-SENSE STATIONARY DATA}
\label{sec:stationary}
If data come from a stationary distribution, cross-covariances depend only on the temporal distance between two observations, $\mtC_{t_1,t_2}=\mtC_{t_1-t_2}=\mtC_\tau$ with $\mtC_{-\tau}=\mtC_\tau^\Tr$, so all blocks on the same diagonal of $\mttC_0$ in \eqref{eq:cov_def} coincide. Since $\mtC_\tau\neq\mtC_\tau^\Tr$ in general, the two diagonals at distance $\tau$ carry different blocks. To maintain (skew-)symmetric Kronecker terms, we set $R=2T-1$, and we split every lagged covariance $\mtC_\tau$ into its symmetric and skew-symmetric parts, $\mtC_\tau^{s}=\tfrac12\big(\mtC_\tau+\mtC_\tau^\Tr\big)$, and
$\mtC_\tau^{a}=\tfrac12\big(\mtC_\tau-\mtC_\tau^\Tr\big)$, such that $\mtC_\tau = \mtC_\tau^{s}+\mtC_\tau^{a}$.
The two terms carry different information. The symmetric part $\mtC_\tau^{s}$ measures how strongly spatial patterns co-move $\tau$ steps apart; the skew part $\mtC_\tau^{a}$ measures which of two patterns moves first, i.e., the lead--lag structure at delay $\tau$, and vanishes whenever the lagged covariance is symmetric.

Then, we use as Kronecker pairs $\mtA_r, \mtB_r$ the matrixes $(\mtI_T,\mthC_0)$ and, for $\tau=1,\dots,T-1$,
\begin{equation}\label{eq:symskew_terms}
    \big(\mtD_\tau+\mtD_\tau^\Tr,\;\mthC_\tau^{s}\big)
    \quad\text{and}\quad
    \big(\mtD_\tau^\Tr-\mtD_\tau,\;\mthC_\tau^{a}\big),
\end{equation}
where $\mtD_\tau\in\mathbb{R}^{T\times T}$ is defined as
\begin{equation}
[\mtD_\tau]_{ij} =
\begin{cases}
1, & \textnormal{if } j - i = \tau,\\
0, & \textnormal{otherwise,}
\end{cases}
\end{equation}
and $\mthC_\tau$ is the sample estimate of $\mtC_\tau$. This is an exact rewriting of the two diagonals at distance $\tau$, since
$(\mtD_\tau+\mtD_\tau^\Tr)\otimes\mtC_\tau^{s}+(\mtD_\tau^\Tr-\mtD_\tau)\otimes\mtC_\tau^{a}=\mtD_\tau\otimes\mtC_\tau^\Tr+\mtD_\tau^\Tr\otimes\mtC_\tau$,
and it requires to estimate only the $T$ spatial matrices $\mthC_0,\dots,\mthC_{T-1}$.  
All temporal matrices in \eqref{eq:symskew_terms} are either symmetric or skew-symmetric, and so are the spatial ones.

\subsection{LOW SEPARATION-RANK COVARIANCE}
\label{sec:low_rank}

Another estimator is proposed in~\cite{tsiligkaridis2013covariance}, which allows to reduce the number of Kronecker terms $R$ even further if the covariance is separable in few Kronecker components.
This relies on a truncated SVD of the permuted sample covariance $\mttR = \mathcal{P}(\mttC_s) \in \mathbb{R}^{T^2 \times N^2}$, where $\mathcal{P}$ is a permutation operator mapping the entries of a matrix of shape $NT \times NT$ to a matrix of shape $T^2 \times N^2$. Given the SVD $\mttR = \mtU \mtSigma \mtV^\Tr$, the first $R$ left and right singular vector pairs $\vcu_r \in \mathbb{R}^{T^2}, \vcv_r \in \mathbb{R}^{N^2}$, $r=1, \dots, R$ are reshaped to be the corresponding terms $\mtA_r, \mtB_r$,
following the property that $\mathcal{P}(\mtA_r \otimes \mtB_r) = \vca_r \vcb_r^\Tr$ where $\vca_r \in \mathbb{R}^{T^2}$, $\vcb_r \in \mathbb{R}^{N^2}$. This allows processing arbitrarily large temporal dependencies $T$ while still maintaining few terms via $R$.

\section{KRONECKER COVARIANCE NEURAL NETWORKS}

Following the VNN principle~\cite{sihag2022covariance,sihag2026learning}, we use spatiotemporal covariance information to build a graph, where the features of samples $\vcx_t$ are node signals, and the edges are covariances at different lags (cf. Figure~\ref{fig:overview}, right).
Then, we propose the spatiotemporal Kronecker coVariance Neural Network (KVNN), a graph convolutional neural network that operates on such graph, represented via the Kronecker decomposition in~\eqref{eq:kron_cov}. We first define the Kronecker coVariance Filter (KVF), which processes an input signal $\vctx$ to produce an output $\vcu$ as: 
\begin{equation}\label{eq:kvf}
    \vcu = \sum_{r=1}^R\mtH_r(\mtA_r, \mtB_r) \vctx = \sum_{r=1}^R \sum_{k=0}^K h_{kr} \mtA_r^k \otimes \mtB_r^k \vctx,
\end{equation}
where we used the property $(\mtA \otimes \mtB)^k = \mtA^k \otimes \mtB^k$.
The filter in~\eqref{eq:kvf} learns different coefficients $\{h_{kr}\}_{k=0}^K$ for each Kronecker term $r$, thus gaining the flexibility to distinctly process correlation information at different time delays.
To show the inherent flexibility of this approach, we compare it with the potential alternative filter design 
$\vcu = \sum_{k=0}^K h_{k} \left(\sum_{r=1}^R \mtA_r \otimes \mtB_r\right)^k \vctx$, adopted by our preliminary work LVNN~\cite{Georgoutsos2025}.
The latter learns a unique set of coefficients $h_k$ for all Kronecker terms, losing expressivity, and the covariance powers $k$ mix the various Kronecker terms $\mtA_r \otimes \mtB_r$, preventing a separate processing of spatiotemporal information at different time delays. 

A KVNN is a sequence of layers followed by non-linear activation functions $\sigma$, where each layer $\ell=1, \dots, L$ contains a filterbank of $F_\textnormal{in} \times F_\textnormal{out}$ KVFs processing the input in parallel:
\begin{equation}\label{eq:kvnn}
    \vctx_f^{(\ell)} = \sigma\left(\sum_{j=1}^{F_\textnormal{in}} \mtH_{fj}^{(\ell)}\vctx_j^{(\ell-1)}\right), \quad f = 1, \dots, F_\textnormal{out}.
\end{equation}
We denote the complete KVNN architecture with $\Phi(\mathcal{H}, \mtC, \vctx)$ for an input signal $\vctx$, a covariance representation $\mtC$ and with $\mathcal{H} = \{ h_{krfj}^{(\ell)} \}_{krfj\ell}$ the learnable coefficients. 

\smallskip
\noindent \textbf{Implementations.}
We define two KVNN implementations which use different Kronecker covariance estimators to achieve different desiderata.
\textbf{KVNN-S} uses the stationary estimator in Section~\ref{sec:stationary}, which allows for theoretical tractability of the frequency response and stability to finite-sample estimations.
\textbf{KVNN-LR}, instead, uses the low-rank estimation in Section~\ref{sec:low_rank}. This decouples the temporal window $T$ from the number of terms to estimate $R$, yielding parameter-efficient estimators even for large temporal dependencies. However, it prevents tractability of the spectral and stability analysis.

\smallskip
\noindent \textbf{Learning sparse weights.} 
In practice, not all blocks in $\mttC$ might be helpful for the downstream task, while they may introduce computational burden.
Thus, we introduce a group sparsity penalty in the training loss such that the KVNN could ignore covariance terms that are not useful for prediction.
Let $\mathcal{H} = \{ h_{krfj} \}_{krfj}$ be the set of all learnable coefficients of a KVNN.
Given a training set $(\mtX_\textnormal{tr}, \mty_\textnormal{tr})$, we train the KVNN to minimize the following objective
\begin{equation}\label{eq:sparse_objective}
    \min_{\mathcal{H}}  \mathcal{L}_\textnormal{task} (\mty_\textnormal{tr}, \Phi(\mathcal{H}, \mtC,\mtX_\textnormal{tr})) + \lambda_g g(\mathcal{H})
\end{equation}
where $\mathcal{L}_\textnormal{task}$ is a task-specific loss (e.g., forecasting MSE) and $g(\mathcal{H})$ is a group-sparsity penalty defined as 
    $g(\mathcal{H}) = \sum_{\ell=1}^L \sum_{r=1}^R \left( \sum_{k,j,f} (h_{krfj}^{(\ell)})^2 \right)^{1/2},$
which promotes sparsity of entire blocks of coefficients $h_{krfj}^{(\ell)}$ for a given value of $r$, thus ignoring some covariance block terms. 
In practice, during training we minimize $g(\mathcal{H})$ by applying a proximal soft-thresholding step after each gradient update. For $\lambda_g > 0$ we additionally prune, before evaluating on validation, a copy of the model in which we set to zero the coefficients $h^{(\ell)}_{krfj}$ whose norm over $(k,f,j)$ at a given $r$ falls below $\alpha$ (set to $0.1$) times the layer's mean group norm.

\begin{figure}[t]
    \centering
    \vspace{-.3cm}
    \includegraphics[width=\linewidth]{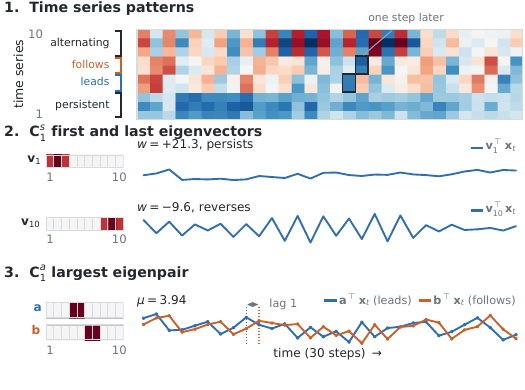}
    \caption{Spectrum of the symmetric and asymmetric lagged covariances on the synthetic time series in (Row 1). (Row 2) The largest eigenvector of $\mtC_\tau^s$ is non-zero on cells with persistent patterns, while the last is non-zero on cells with highly alternating patterns at delay $\tau$. 
    (Row 3) The eigenvector pair $\vca,\vcb$ of $\mtC_\tau^a$ is non-zero for cells that have correlated lead-follow patterns at delay $\tau$. 
    \vspace{-.5cm}}
    \label{fig:spectrum}
\end{figure}

\section{THEORETICAL ANALYSIS}\label{sec:theory}
 
We present the spectral analysis of KVNN-S (along with its filter KVF-S), and its stability to finite-sample errors. Proofs are collected in the Appendix.
 
\subsection{SPECTRAL ANALYSIS}\label{sec:spectral}
 
A unified spectral analysis of the KVF-S is challenging because the Kronecker terms have different eigenvectors. However, since the KVF-S learns different coefficients on distinct Kronecker terms, we analyze each term separately. This characterizes the filter expressivity and processing of spatiotemporal information and connects it to classical spectral tools.
 
For a generic $r$, let $\mtA_r=\mtU_r\mtLambda_r\mtU_r^\Her$ and $\mtB_r=\mtV_r\mtW_r\mtV_r^\Her$ be the unitary eigendecompositions of the two factors. When $\mtA_r,\mtB_r$ are symmetric, $\mtU_r,\mtV_r$ are real orthogonal and $\mtLambda_r,\mtW_r$ are real; when they are skew-symmetric, $\mtU_r,\mtV_r$ are complex unitary and $\mtLambda_r,\mtW_r$ are purely imaginary. In both cases the products $\lambda_{r,m}w_{r,n}$ are real, and the term $\mtA_r\otimes\mtB_r$ is a real symmetric matrix. 
We define the $r$-th \textit{spatiotemporal covariance Fourier transform} for a KVF-S and a signal $\vctx$ as $\vchx = (\mtU_r \otimes \mtV_r)^\Her \vctx$.
For each $r$ (or equivalently $\tau$), this eigenspectrum has an interpretable meaning.
The temporal factors $\mtD_\tau+\mtD_\tau^\Tr$ and $\mtD_\tau^\Tr-\mtD_\tau$ are the $\tau$-step undirected line-graph shift and its skew counterpart. For $\tau=1$ their eigenvectors are the discrete sinusoids that graph-time signal processing adopts as temporal frequencies~\cite{sandryhaila2014discrete}. 
The spatial factor $\mtC_\tau^{s}$ has as eigenvectors spatial patterns and as eigenvalues their lag-$\tau$ autocovariance, i.e., patterns are ranked by how strongly they carry over $\tau$ steps.
These are the basis of time-lagged component analysis~\cite{belouchrani1997blind,molgedey1994separation}. 
The spatial factor $\mtC_\tau^{a}$ has eigenvalues $w=\pm i\mu$ attached to pairs of orthogonal patterns $(\vca,\vcb)$, where $\mu$ measures their co-variation at delay $\tau$. 
Figure~\ref{fig:spectrum} shows an example. The first and last eigenvectors of $\mtC_1^s$ identify, respectively, persistent signals and patterns alternating every $\tau=1$ steps. The largest eigenpair of $\mtC_1^a$, instead, identifies a pattern where one signal leads another one by $\tau=1$ step.  

\begin{figure}[t]
    \centering
    \vspace{-.7cm}
    \includegraphics[width=\linewidth]{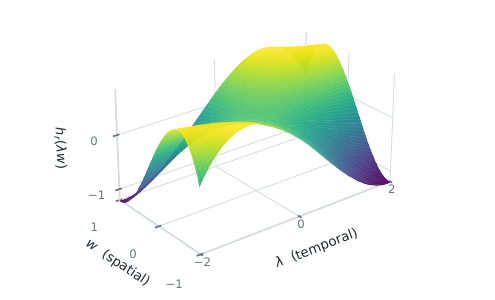}
    \caption{Frequency response $h_r(\lambda w)$ of the KVF-S on the $r$-th Kronecker term. \vspace{-.5cm}}
    \label{fig:freq_response}
\end{figure}

We now analyze the KVF-S in the $r$-th eigenspectrum.
By properties of the Kronecker product, we get
\begin{align}
\vcu_r &= \sum_{k=0}^K h_{kr} (\mtU_r \otimes \mtV_r) (\mtLambda_r^k \otimes \mtW_r^k)(\mtU_r \otimes \mtV_r)^\Her \vctx \\
&= (\mtU_r \otimes \mtV_r) \sum_{k=0}^K h_{kr} (\mtLambda_r^k \otimes \mtW_r^k) \vchx.
\end{align}
For the $i$-th entry of $\vcu$, indexed by a temporal--spatial pair $(m,n)$, this corresponds to
\begin{align}\label{eq:freq_response}
    [\vcu_r]_i = \sum_{k=0}^K h_{kr}\,\lambda_{r,m}^kw_{r,n}^k\, [\vchx]_i = h_r(\lambda_{r,m}w_{r,n})\,[\vchx]_i .
\end{align}
That is, the KVF-S learns a polynomial frequency response $h_r(\gamma)$ in the variable $\gamma=\lambda w$, the product of a temporal and a spatial eigenvalue of the $r$-th Kronecker term (cf. Figure~\ref{fig:freq_response}).

\subsection{STABILITY ANALYSIS}\label{sec:stability}
 
We analyze the stability of the KVF-S to finite-sample estimation errors. The temporal matrices $\mtA_r$ are fixed, so the finite-sample estimation error lies only in the spatial matrices. We present the Lipschitz filter assumption, then state our stability result.
 
\begin{assumption}\label{as:lipschitz}
The frequency response $h_r$ in~\eqref{eq:freq_response} is Lipschitz with constant $P$ on an interval $\mathcal{I}_r\subset\mathbb{R}$ containing the eigenvalues of both $\mtA_r\otimes\mtB_r$ and $\mtA_r\otimes\mthB_r$, i.e., $|h_r(\gamma)-h_r(\gamma')|\leq P|\gamma-\gamma'|$ for all $\gamma,\gamma'\in\mathcal{I}_r$.
\end{assumption}

\begin{theorem}\label{thm:kvf_stability}
Consider a KVF satisfying Assumption~\ref{as:lipschitz} and operating on the stationary Kronecker terms \eqref{eq:symskew_terms}, whose spatial matrices are estimated from $M>NT$ independent, zero-mean, sub-Gaussian observations of a wide-sense stationary process, stacked as in~\eqref{eq:sample_cov}. Let $\vcu_0$ be the KVF output on the true lagged covariances and $\vcu$ the output on their sample estimates. Then
\begin{equation}\label{eq:stability_bound}
    \|\vcu_0-\vcu\|_2
    = \mathcal{O}_P\!\left(\frac{T^{3/2}PN}{\sqrt{M}}\right)\|\vctx\|_2 ,
\end{equation}
where $\mathcal{O}_P$ indicates that the bound holds with probability increasing to $1$ as $M\to\infty$. 
\end{theorem}
 
The stability of the KVF depends on its Lipschitz constant $P$: a larger $P$ allows sharper variations of the response across spatiotemporal modes, which in turn can amplify errors in the estimated spectrum. The filter converges to the ideal one at rate $\mathcal{O}(M^{-1/2})$, like standard covariance estimators, and the bound grows with $N$ and $T$ as these increase the number of covariance entries to estimate.  
This result can be extended to the complete KVNN by applying~\cite[Theorem 4]{gama2020stability}. 
Notably, the KVNN stability bound does not increase if covariance eigenvalues are close. This happens, instead, for ST-PCA, which requires an explicit eigendecomposition (cf. \cite[Proposition 1]{cavallo2024stvnn}). The advantage of KVNN lies in its spectral processing via polynomial functions bounded by the Lipschitz constant, which ensure that eigenvalue perturbations result in bounded output differences. This analysis extends analogous observations in the parallel between VNNs and PCA~\cite{sihag2022covariance}.
Finally, while independence of the windows is an idealization for stationary data, most of our stability proof does not use it, and it can be replaced by a finite-memory or spectral-density condition on the process at the cost of constants (cf.~\cite{basu2015regularized}).

\begin{table*}[t]
\centering
\vspace{-.5cm}
\caption{Test MAE ($\times 10$, mean $\pm$ std, lower is better) by prediction horizon $h$. 
\textbf{Best} and \underline{\textit{second best}} results highlighted.}
\label{tab:main_free}
\footnotesize
\begin{tabular}{lcccccc|ccc}
\toprule
Dataset & $h$ & Ref. & LSTM & ST-PCA & VNN & STVNN & LVNN (Ours) & KVNN-S (Ours) & KVNN-LR (Ours) \\
\midrule
\multirow{3}{*}{LargeCap} & 1 & 8.02 & 7.98\,$\pm$\,0.04 & 8.00\,$\pm$\,0.03 & 7.84\,$\pm$\,0.08 & 7.63\,$\pm$\,0.10 & \underline{\textit{7.42\,$\pm$\,0.08}} & 7.57\,$\pm$\,0.04 & \textbf{7.40\,$\pm$\,0.04} \\
 & 2 & 8.01 & 8.03\,$\pm$\,0.04 & 8.04\,$\pm$\,0.02 & 7.88\,$\pm$\,0.03 & 7.73\,$\pm$\,0.10 & \textbf{7.51\,$\pm$\,0.07} & 7.64\,$\pm$\,0.04 & \underline{\textit{7.53\,$\pm$\,0.04}} \\
 & 3 & 8.01 & 8.05\,$\pm$\,0.04 & 8.06\,$\pm$\,0.01 & 7.96\,$\pm$\,0.05 & 7.78\,$\pm$\,0.08 & \underline{\textit{7.58\,$\pm$\,0.07}} & 7.71\,$\pm$\,0.08 & \textbf{7.57\,$\pm$\,0.03} \\
\midrule
\multirow{3}{*}{Sectors} & 1 & 2.59 & 1.58\,$\pm$\,0.02 & 1.75\,$\pm$\,0.03 & 2.07\,$\pm$\,0.24 & 1.64\,$\pm$\,0.10 & \underline{\textit{1.53\,$\pm$\,0.03}} & 1.58\,$\pm$\,0.07 & \textbf{1.52\,$\pm$\,0.02} \\
 & 2 & 2.59 & 1.60\,$\pm$\,0.02 & 1.73\,$\pm$\,0.00 & 2.15\,$\pm$\,0.24 & 1.63\,$\pm$\,0.05 & \underline{\textit{1.58\,$\pm$\,0.04}} & 1.61\,$\pm$\,0.04 & \textbf{1.55\,$\pm$\,0.00} \\
 & 3 & 2.59 & \underline{\textit{1.63\,$\pm$\,0.03}} & 1.82\,$\pm$\,0.01 & 2.18\,$\pm$\,0.16 & 1.70\,$\pm$\,0.10 & 1.65\,$\pm$\,0.08 & 1.66\,$\pm$\,0.05 & \textbf{1.61\,$\pm$\,0.02} \\
\midrule
\multirow{3}{*}{NN5} & 1 & 7.31 & \textbf{5.42\,$\pm$\,0.07} & 5.81\,$\pm$\,0.08 & 7.77\,$\pm$\,0.08 & 5.90\,$\pm$\,0.05 & 6.05\,$\pm$\,0.20 & \underline{\textit{5.65\,$\pm$\,0.09}} & 5.73\,$\pm$\,0.06 \\
 & 2 & 7.29 & \textbf{5.61\,$\pm$\,0.04} & 6.04\,$\pm$\,0.06 & 7.68\,$\pm$\,0.05 & 6.20\,$\pm$\,0.02 & 6.21\,$\pm$\,0.21 & \underline{\textit{5.81\,$\pm$\,0.04}} & 5.95\,$\pm$\,0.06 \\
 & 3 & 7.27 & \textbf{5.87\,$\pm$\,0.06} & 6.14\,$\pm$\,0.06 & 7.63\,$\pm$\,0.30 & 6.26\,$\pm$\,0.01 & 6.21\,$\pm$\,0.03 & \underline{\textit{5.94\,$\pm$\,0.04}} & 6.13\,$\pm$\,0.18 \\
\midrule
\multirow{3}{*}{CzeLan} & 1 & 5.42 & 4.45\,$\pm$\,0.07 & 4.52\,$\pm$\,0.04 & 4.26\,$\pm$\,0.01 & 4.47\,$\pm$\,0.01 & 4.32\,$\pm$\,0.00 & \textbf{4.21\,$\pm$\,0.02} & \underline{\textit{4.22\,$\pm$\,0.02}} \\
 & 3 & 5.41 & 4.61\,$\pm$\,0.06 & 4.77\,$\pm$\,0.06 & 4.58\,$\pm$\,0.01 & 4.66\,$\pm$\,0.01 & \textbf{4.46\,$\pm$\,0.02} & 4.53\,$\pm$\,0.01 & \underline{\textit{4.50\,$\pm$\,0.00}} \\
 & 6 & 5.41 & 4.65\,$\pm$\,0.02 & 4.81\,$\pm$\,0.04 & 4.73\,$\pm$\,0.01 & 4.80\,$\pm$\,0.03 & \textbf{4.55\,$\pm$\,0.01} & 4.66\,$\pm$\,0.03 & \underline{\textit{4.60\,$\pm$\,0.01}} \\
\midrule
\multirow{3}{*}{AQWan} & 1 & 6.03 & 5.62\,$\pm$\,0.05 & 5.75\,$\pm$\,0.05 & 5.44\,$\pm$\,0.03 & 5.43\,$\pm$\,0.01 & 5.34\,$\pm$\,0.01 & \underline{\textit{5.33\,$\pm$\,0.04}} & \textbf{5.26\,$\pm$\,0.04} \\
 & 3 & 6.03 & 5.80\,$\pm$\,0.05 & 5.81\,$\pm$\,0.05 & 5.92\,$\pm$\,0.01 & 5.61\,$\pm$\,0.01 & 5.57\,$\pm$\,0.01 & \underline{\textit{5.56\,$\pm$\,0.01}} & \textbf{5.54\,$\pm$\,0.05} \\
 & 6 & 6.03 & 5.80\,$\pm$\,0.03 & 5.77\,$\pm$\,0.00 & 5.95\,$\pm$\,0.01 & 5.65\,$\pm$\,0.03 & \underline{\textit{5.59\,$\pm$\,0.02}} & \textbf{5.58\,$\pm$\,0.02} & 5.59\,$\pm$\,0.01 \\
\bottomrule
\end{tabular}
\end{table*}

\begin{table}[t]
\centering
\vspace{-.2cm}
\caption{Number of learnable parameters for each model in the configuration reported in Table~\ref{tab:main_free}.}
\label{tab:params_free}
\footnotesize
\setlength{\tabcolsep}{4pt}
\begin{tabular}{lrrrrr}
\toprule
Model & LargeCap & Sectors & NN5 & CzeLan & AQWan \\
\midrule
LSTM & 31\,788 & 7\,553 & 287\,123 & 358\,119 & 358\,119 \\
ST-PCA & 8\,124 & 42\,401 & 13\,107 & 5\,351 & 142\,375 \\
VNN & 71\,945 & 104\,841 & 71\,945 & 406\,153 & 137\,737 \\
STVNN & 22\,025 & 169\,225 & 1\,641 & 13\,577 & 51\,593 \\
LVNN & 71\,965 & 10\,397 & 137\,744 & 274\,849 & 274\,849 \\
KVNN-S & 42\,675 & 22\,035 & 87\,511 & 26\,031 & 101\,249 \\
KVNN-LR & 9\,405 & 1\,309 & 53\,069 & 15\,637 & 19\,809 \\
\bottomrule
\end{tabular}
\vspace{-.3cm}
\end{table}

\section{NUMERICAL EXPERIMENTS}

Our numerical experiments have the following objectives.
\textbf{(O1)} Demonstrate that KVNNs are effective at real-data forecasting and their design choices are motivated by performance improvements w.r.t. relevant baselines.
\textbf{(O2)} Show that our stability analysis translates into consistent KVNNs' performance under finite-sample covariance estimation errors.
\textbf{(O3)} Support the group-sparsity penalty formulation showing that it identifies meaningful temporal lags.
\textbf{(O4)} Analyze the impact of hyperparameters on performance.
Our code is available online\footnote{\url{https://github.com/andrea-cavallo-98/KVNN}}.

\subsection{FORECASTING (O1)}

\smallskip \noindent \textbf{Datasets.}
We use five real-world multivariate time series.
Three come from the TFB forecasting benchmark~\cite{qiu2024tfb}.
\textbf{NN5} contains $N=111$ daily series of cash withdrawals at UK ATMs over $791$ days, with a weekly seasonality.
\textbf{CzeLan} contains $N=11$ hourly ecological channels recorded at a Czech forest site over $19{,}934$ steps.
\textbf{AQWan} contains $N=11$ hourly air-quality channels from a Beijing monitoring station over $35{,}064$ steps.
The remaining two are variance forecasting tasks that we construct from publicly available daily price data. The input collects the daily log returns of $N$ assets, and the target is the vector of their within-day variances on future days, computed with the Garman--Klass estimator~\cite{garman1980estimation}. 
\textbf{Sectors} comprises the $N=9$ original Select Sector SPDR exchange-traded funds over $6{,}961$ trading days (1999--2026).
\textbf{LargeCap} comprises $N=50$ large-capitalization U.S.\ stocks over $1{,}425$ days (2021--2026).

We split every series chronologically into $60\%$ training, $20\%$ validation and $20\%$ test, and standardized with training statistics.
We use CzeLan and AQWan in first differences, $\vcx_t \leftarrow \vcx_t - \vcx_{t-1}$, since in levels they are dominated by persistence, we truncate them and to their first $10{,}000$ steps.
We extract windows of $T$ consecutive samples at unit stride.

\smallskip \noindent \textbf{Models.}
We evaluate three models of the proposed family:
\textbf{KVNN-S}, which uses the stationary Kronecker estimator; 
\textbf{KVNN-LR}, which uses the low-rank estimator;
and \textbf{LVNN}~\cite{Georgoutsos2025}, which operates on the full $\mttC_s$ with coefficients shared across lags.
Each baseline isolates one design choice of KVNNs:
\textbf{VNN}~\cite{sihag2022covariance} ignores temporal information;
\textbf{STVNN}~\cite{cavallo2024stvnn} uses a temporal window but neglects delayed correlations;
\textbf{ST-PCA} projects the windowed input onto the top $q$ eigenvectors of $\mttC_s$ before a forecasting head, i.e., it processes spatiotemporal covariance disjointly from the task;
and an \textbf{LSTM}~\cite{hochreiter1997long} uses no covariance information.
Finally, \textbf{Ref.}\ is the best of naive predictors: the training mean and, where available, persistence and seasonal naive.

\smallskip \noindent \textbf{Experimental setup.}
We forecast all horizons jointly, $h \in \{1,2,3\}$ for NN5, Sectors and LargeCap and $h \in \{1,3,6\}$ for CzeLan and AQWan, and report the test MAE on standardized targets as mean $\pm$ standard deviation over $3$ seeds.
We optimize hyperparameters on validation set via a grid search among the following values:
$T$ (three dataset-specific values), depth $L \in \{1,2,3\}$ and filter order $K \in \{1,2,3\}$; in addition $\lambda_g \in \{0, 0.3, 1\}$ for KVNN-S and KVNN-LR, $R \in \{1,3,8\}$ for KVNN-LR, $q \in \{8,32,128\}$ for ST-PCA, and a width ladder for each baseline. 
For KVNN-S and KVNN-LR, we fix $F=32$, while for each baseline we increase the feature size until its largest configuration is at least as large as KVNN's largest.
All remaining settings are shared by every model and configuration: Adam with learning rate $10^{-2}$, batch size $128$, dropout $0.1$, a 2-layer MLP head, a softmax temporal readout, an autoregressive skip initialized to exact persistence, Kronecker blocks normalized to unit spectral norm, and at most $600$ epochs with early stopping on validation MAE after $40$ epochs without improvement.  
Table~\ref{tab:main_free} reports the results, and Table~\ref{tab:params_free} the number of trainable parameters of the selected configurations.

\smallskip \noindent \textbf{Discussion.}
At least one member of the KVNN family (LVNN, KVNN-S, KVNN-LR) ranks in the top two in all $15$ dataset--horizon settings; the family occupies both first and second place in $11$ of them and provides the best model in $12$.
The three covariance-based baselines highlight the value of each design choice:
(i) the gains over ST-PCA show that end-to-end, task-aware covariance processing outperforms a disjoint project-then-forecast pipeline;
(ii) the gains over VNN show that temporal information is essential;
(iii) the gains over STVNN show that delayed cross-covariances carry information that the snapshot covariance alone misses.

Among the proposed models, KVNN-LR is the most consistent: it is the best model in $7$ of $15$ settings and does so with fewer parameters than KVNN-S and LVNN on every dataset (Table~\ref{tab:params_free}), e.g., $1.3$k against $22$k and $10$k on Sectors.
This indicates that a low-rank Kronecker representation captures the relevant delayed structure more parsimoniously than the explicit lag-by-lag parametrization of KVNN-S.
LVNN is competitive but less consistent, since it cannot weigh different lags differently.
 
Every KVNN model improves on Ref., confirming that the tasks are non-trivial.
The LSTM, which uses no covariance information, is the least consistent learned model: it barely improves on Ref.\ on LargeCap and is among the weakest models on AQWan, yet it is the best model on NN5, where it outperforms KVNN-S by a small margin, but uses more than three times the parameter count.
This supports the view that the covariance provides a useful inductive bias: it substantially reduces the number of trainable parameters at the cost of, at most, a minor loss in accuracy.

\begin{figure*}
\vspace{-.5cm}
    \centering
    \includegraphics[width=\linewidth]{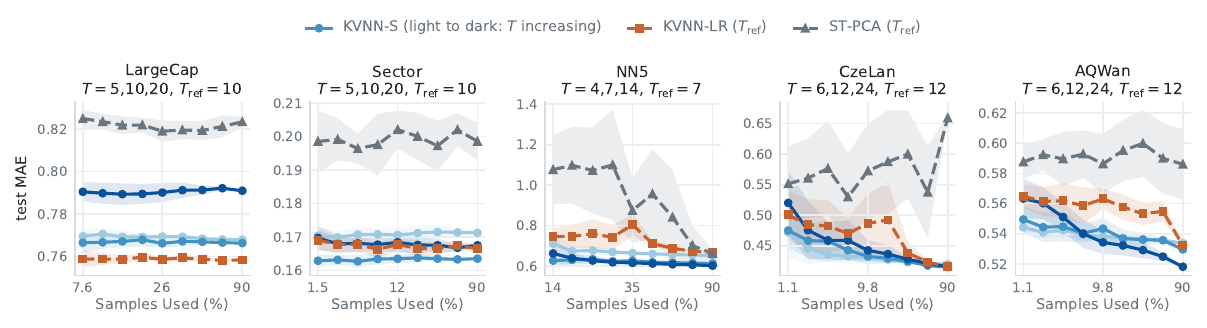}
    \caption{Stability to finite-sample covariance estimation errors on the five real-world datasets.  \vspace{-.5cm}}
    \label{fig:stability}
\end{figure*}

\begin{figure*}[t]
    \centering
    \includegraphics[width=\linewidth]{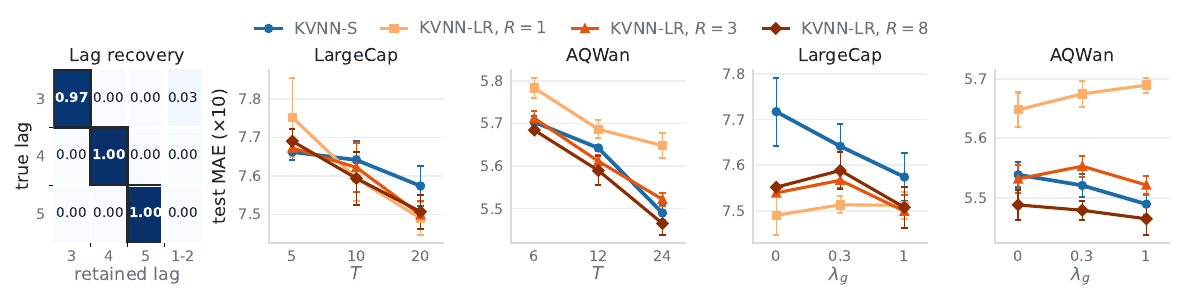}
    \caption{(Left) Share of the retained coefficient mass per lag ($\lambda_g=2$), for a process whose only non-zero lagged covariance is at $\tau=q$ (rows).
    The group-sparsity penalty concentrates the surviving coefficients on the informative lag. (Middle) Impact of hyperparameters $T,R$ ($R$ in the legend). (Right) Impact of hyperparameters $\lambda_g,R$. \vspace{-.5cm}}
    \label{fig:lag_matrix}
\end{figure*}

\subsection{STABILITY (O2)}

We test empirically the finite-sample stability predicted by Theorem~\ref{thm:kvf_stability}.
For each dataset, KVNN-S, KVNN-LR and ST-PCA are trained once with the covariance operator estimated from the full training series and then frozen.
We then re-estimate the covariance from a contiguous block of $M$ consecutive training samples with a uniformly random start, substitute it into the frozen model, and measure the test MAE.
Since nothing but the covariance operator changes, the resulting degradation isolates the effect of estimation error on the operator from differences in what was learned.
For each $M$, we recompute the covariance $20$ times with different data subsets and report the mean results $\pm 1$ standard deviation.
All models share one configuration: a single layer with $F=32$, $K=1$, an MLP head of sizes $(32,16)$, dropout $0.1$, Adam with learning rate $10^{-2}$, batch size $128$, and at most $300$ epochs with early stopping after $40$.
We set $\lambda_g = 0$ so that no Kronecker term is pruned, which would alter the operator under study.
KVNN-S ($R=2T-1$) is shown for the three values of $T$ in each dataset's grid; KVNN-LR ($R=8$) and ST-PCA ($q=32$) are shown at the intermediate value $T_\textnormal{ref}$.

Figure~\ref{fig:stability} shows that KVNNs are markedly more stable than ST-PCA.
The test MAE of KVNNs stays close to its full-sample value down to very small fractions of the training data and decreases monotonically as $M$ grows, whereas ST-PCA is uniformly worse, exhibits a larger variance across blocks, and on NN5 degrades sharply when $M$ is small.
Within KVNN-S, a larger $T$ leads to a larger gap between the small-$M$ and full-sample MAE, consistent with the dependence on $T$ in Theorem~\ref{thm:kvf_stability}, although on LargeCap and Sectors the curves are essentially flat and the effect is negligible.
KVNN-LR is less stable than KVNN-S due to the explicit SVD required to estimate the Kronecker terms. Neveretheless, it provides more consistent performance compared to ST-PCA.

\subsection{LAG RECOVERY (O3)}

We verify that the penalty $g(\mathcal{H})$ removes the Kronecker terms that carry no correlation in a controlled setting.
We generate a moving-average process $\vcx_t = \boldsymbol{\varepsilon}_t + 0.9\,\mathbf{Q}\,\boldsymbol{\varepsilon}_{t-q}$ with $\boldsymbol{\varepsilon}_t \sim \mathcal{N}(\mathbf{0}, \mathbf{I}_N)$, $\mathbf{Q}$ a symmetric orthogonal matrix
and $N=8$, so that every $\mtC_\tau$ is symmetric and each lag is carried by a single Kronecker term.
By construction, $\mtC_\tau = \mathbf{0}$ for $\tau \notin \{0, q\}$, so the corresponding blocks $\mtB_r$, estimated from $\mthC_\tau$, contain pure estimation noise.
We forecast $\vctx_{t+q}$ from a window of $T=6$ steps and read out only its last position, so that every cross-time relation must pass through a Kronecker term.
We train a single-layer KVNN-S ($F=16$, $K=1$, blocks normalized to unit spectral norm) on $M = 1.2 \cdot 10^4$ windows for $60$ epochs with Adam ($\eta = 10^{-2}$, batch size $128$).

Figure~\ref{fig:lag_matrix} (left) reports, for each true lag $q$, the share of the retained coefficient mass carried by each lag at $\lambda_g = 2$, averaged over three seeds.
The penalty concentrates $0.97$, $1.00$ and $1.00$ of the mass on the true lag for $q = 3, 4, 5$, respectively, leaving at most $0.03$ on any wrong lag. Lags 1-2 are grouped in the same column as they are never the ground truth.

\subsection{IMPACT OF HYPERPARAMETERS (O4)}

Figure~\ref{fig:lag_matrix} (middle and right) shows the effect of $T$, $R$ and $\lambda_g$ on LargeCap and AQWan.
Increasing $T$ improves all KVNN variants on both datasets, as the model accesses longer temporal dependencies.
A larger $T$ does not require a larger $R$ in KVNN-LR. Often, $R=1$ is insufficient, but $R=3$ already matches $R=8$ and KVNN-S at all $T$, so a handful of Kronecker terms suffices to represent windows of up to $24$ steps.
The effect of $\lambda_g$ is model- and dataset-dependent. KVNN-S benefits from stronger sparsification on LargeCap, KVNN-LR with $R \in \{3,8\}$ is largely insensitive to it, and KVNN-LR with $R=1$ slightly degrades on AQWan, as pruning any of its few terms removes useful information.
Overall, $\lambda_g$ is not a critical hyperparameter for accuracy, and its main benefit is the reduced number of active Kronecker terms.


\section{CONCLUSION}
We introduced the spatiotemporal Kronecker coVariance Neural Network (KVNN), an architecture designed to efficiently and flexibly process the joint spatiotemporal correlations inherent in multivariate time series via a sum-of-Kronecker representation.
We provided a spectral analysis and proved that, under stationary assumptions, the KVF maintains strong stability bounds against finite-sample estimation errors.
Our empirical evaluations on real-world forecasting tasks validated the forecasting capability of KVNNs on five datasets.

\section*{APPENDIX}
 
We first provide a useful Lemma, then prove Theorem~\ref{thm:kvf_stability}.

\begin{lemma}\label{lem:lipschitz_hermitian}
Let $\mtM,\mthM\in\mathbb{C}^{N\times N}$ be Hermitian with eigenvalues $\{\alpha_i\}$ and $\{\beta_j\}$, and let $h$ be a real polynomial with $|h(\alpha_i)-h(\beta_j)|\leq P|\alpha_i-\beta_j|$ for all $i,j$. Then
\begin{align}
    \|h(\mtM)-h(\mthM)\|_2 &\leq  P\sqrt{N}\,\|\mtM-\mthM\|_2 .
\end{align}
\end{lemma}
\begin{myproof}
Write $\mtM=\sum_i\alpha_i\vcp_i\vcp_i^\Her$ and $\mthM=\sum_j\beta_j\vcq_j\vcq_j^\Her$ with orthonormal bases $\{\vcp_i\}$, $\{\vcq_j\}$. Since $\vcp_i^\Her h(\mtM)$ and $h(\mthM)\vcq_j$ act through the respective eigenvalues, we get
\begin{align}
    \vcp_i^\Her\big(h(\mtM)-h(\mthM)\big)\vcq_j &= \big(h(\alpha_i)-h(\beta_j)\big)\,\vcp_i^\Her\vcq_j.
\end{align}
By setting $h(x)=x$:
\begin{align}
    \vcp_i^\Her\big(\mtM-\mthM\big)\vcq_j &= \big(\alpha_i-\beta_j\big)\,\vcp_i^\Her\vcq_j .
\end{align}

Hence $|\vcp_i^\Her(h(\mtM)-h(\mthM))\vcq_j|\leq P\,|\vcp_i^\Her(\mtM-\mthM)\vcq_j|$ for all $i,j$ (both sides vanish if $\alpha_i=\beta_j$). For any $\mtX$ and unitary $\mtP=[\vcp_1,\dots,\vcp_N]$, $\mtQ=[\vcq_1,\dots,\vcq_N]$ we have $\|\mtX\|_F^2=\|\mtP^\Her\mtX\mtQ\|_F^2=\sum_{i,j}|\vcp_i^\Her\mtX\vcq_j|^2$; summing the squared inequalities over $i,j$ gives $\|h(\mtM)-h(\mthM)\|_F\leq P\|\mtM-\mthM\|_F$. Finally, the norm inequalities $\|\cdot\|_2\leq\|\cdot\|_F\leq\sqrt{N}\|\cdot\|_2$ complete the result.
\end{myproof}
 
\subsection*{PROOF OF THEOREM~\ref{thm:kvf_stability}}
Let $\mtH_r(\mtM)=\sum_{k=0}^K h_{kr}\mtM^k$ and $\mtE_\tau=\mthC_\tau-\mtC_\tau$ for $\tau=0,\dots,T-1$. By triangle inequality, we get
\begin{align}
    \|\vcu_0&-\vcu\|_2
    =\Big\|\sum_r\big[\mtH_r(\mtA_r\otimes\mtB_r)-\mtH_r(\mtA_r\otimes\mthB_r)\big]\vctx\Big\|_2 \nonumber
    \\&\leq\sum_r\big\|\mtH_r(\mtA_r\otimes\mtB_r)-\mtH_r(\mtA_r\otimes\mthB_r)\big\|_2\,\|\vctx\|_2 . \label{eq:pf_triangle}
\end{align} 
Then, for a fixed $r$ (we omit the subscript $r$ from eigenvectors for simplicity), using the mixed-product property yields
\begin{equation}
    \mtA_r\otimes\mtB_r=(\mtU\otimes\mtI_N)(\mtLambda\otimes\mtB_r)(\mtU\otimes\mtI_N)^\Her ,
\end{equation}
and since $\mtU\otimes\mtI_N$ is unitary, $(\mtA_r\otimes\mtB_r)^k=(\mtU\otimes\mtI_N)(\mtLambda\otimes\mtB_r)^k(\mtU\otimes\mtI_N)^\Her$ for all $k$, so that
$\mtH_r(\mtA_r\otimes\mtB_r)=(\mtU\otimes\mtI_N)\mtH_r(\mtLambda\otimes\mtB_r)(\mtU\otimes\mtI_N)^\Her$, and likewise for $\mthB_r$. By unitary invariance of the spectral norm,
\begin{align}
    \big\|\mtH_r(\mtA_r\otimes\mtB_r)-\mtH_r(\mtA_r\otimes\mthB_r)\big\|_2= \nonumber \\
    \big\|\mtH_r(\mtLambda\otimes\mtB_r)-\mtH_r(\mtLambda\otimes\mthB_r)\big\|_2 = \nonumber \\
    \max_{i}\big\|\mtH_r(\lambda_i\mtB_r)-\mtH_r(\lambda_i\mthB_r)\big\|_2,\label{eq:pf_blocks}
\end{align}
where the last step uses the fact that the matrix $\mtLambda\otimes\mtB_r$ is block-diagonal with $N\times N$ blocks $\lambda_i\mtB_r$, a polynomial acts block-wise, and the spectral norm of a block-diagonal matrix is the maximum over its blocks. 

If $\mtA_r,\mtB_r$ are symmetric, $\lambda_i$ is real and $\lambda_i\mtB_r$, $\lambda_i\mthB_r$ are real symmetric. If $\mtA_r,\mtB_r$ are skew-symmetric, $\lambda_i=i\theta_i$ with $\theta_i\in\mathbb{R}$, and $(\lambda_i\mtB_r)^\Her=-i\theta_i\mtB_r^\Tr=i\theta_i\mtB_r=\lambda_i\mtB_r$, so the blocks are Hermitian; the same holds for $\lambda_i\mthB_r$. In both cases the eigenvalues of $\lambda_i\mtB_r$ and $\lambda_i\mthB_r$ are of the form $\lambda_iw$ with $w$ an eigenvalue of $\mtB_r$ or $\mthB_r$, i.e., they are eigenvalues of $\mtA_r\otimes\mtB_r$ or $\mtA_r\otimes\mthB_r$. Assumption~\ref{as:lipschitz} and Lemma~\ref{lem:lipschitz_hermitian} give
\begin{align}
    \big\|\mtH_r(\lambda_i\mtB_r)&-\mtH_r(\lambda_i\mthB_r)\big\|_2
    \leq P\sqrt{N}\,\|\lambda_i(\mtB_r-\mthB_r)\|_2 \nonumber
    \\&= P\sqrt{N}\,|\lambda_i|\,\|\mtB_r-\mthB_r\|_2 .
\end{align}
Taking the maximum over $i$ in \eqref{eq:pf_blocks} yields
\begin{equation}\label{eq:pf_term}
    \big\|\mtH_r(\mtA_r\otimes\mtB_r)-\mtH_r(\mtA_r\otimes\mthB_r)\big\|_2
    \leq P\sqrt{N}\,\|\mtA_r\|_2\,\|\mtB_r-\mthB_r\|_2 .
\end{equation}
 
For $\tau=0$, $\|\mtI_T\|_2=1$ and $\mtB_r-\mthB_r=-\mtE_0$. 
For $\tau\ge1$, every row and every column of $\mathbf{D}_\tau$ contains at most one non-zero entry, equal to one, so $\mathbf{D}_\tau$ is a partial permutation matrix and $\|\mathbf{D}_\tau\|_2=\|\mathbf{D}_\tau^\top\|_2=1$. By the triangle inequality, $\|\mathbf{D}_\tau\pm\mathbf{D}_\tau^\top\|_2\leq 2$.
Moreover,
\begin{align}
    \|\mtC_\tau^{s}-\mthC_\tau^{s}\|_2=\tfrac12\|\mtE_\tau+\mtE_\tau^\Tr\|_2\leq\|\mtE_\tau\|_2 ,
    \\
    \|\mtC_\tau^{a}-\mthC_\tau^{a}\|_2=\tfrac12\|\mtE_\tau-\mtE_\tau^\Tr\|_2\leq\|\mtE_\tau\|_2 .
\end{align}
Substituting in \eqref{eq:pf_term} and summing over the $2T-1$ terms in \eqref{eq:pf_triangle},
\begin{align}
    \|\vcu_0-\vcu\|_2
    &\leq P\sqrt{N}\Big(\|\mtE_0\|_2+2\sum_{\tau=1}^{T-1}\big(\|\mtE_\tau\|_2+\|\mtE_\tau\|_2\big)\Big)\|\vctx\|_2 \nonumber \\
    & \leq 4P\sqrt{N}\Big(\sum_{\tau=0}^{T-1}\|\mtE_\tau\|_2\Big)\|\vctx\|_2 \nonumber \\
    &\leq 4PT\sqrt{N}\,\max_\tau\|\mtE_\tau\|_2\,\|\vctx\|_2 . \label{eq:pf_deterministic}
\end{align}
This deterministic bound holds for any realization of the estimation errors.

Since $\mthC_\tau$ is the average of the $T-\tau$ blocks on the $\tau$-th block diagonal of $\mttC_s$, the error $\mtE_\tau$ is a convex combination of sub-matrices of $\mttC_s-\mttC_0$, and hence $\max_\tau\|\mtE_\tau\|_2\leq\|\mttC_s-\mttC_0\|_2$. The windows $\vctx_t$ are $M$ independent zero-mean sub-Gaussian vectors in $\mathbb{R}^{NT}$ with covariance $\mttC_0$, so by \cite[Theorem~4.7.1]{vershynin2018high} there is a universal constant $C>0$ such that for every $u\ge0$, with probability at least $1-2e^{-u}$,
\begin{align}
    \|\mttC_s-\mttC_0\|_2&\leq C\kappa^2\|\mttC_0\|_2\left(\sqrt{\frac{NT+u}{M}}+\frac{NT+u}{M}\right)\nonumber
    \\&\leq c\sqrt{\frac{NT+u}{M}},
\end{align}
where $\kappa$ is the sub-Gaussian norm of $\vctx$, $c=2C\kappa^2\|\mttC_0\|_2$, and the last step uses $M\ge NT+u$. Inserting this into \eqref{eq:pf_deterministic} with $u=\log(2/\delta)$ and absorbing $\delta$ and constants gives \eqref{eq:stability_bound}. \hfill$\blacksquare$

\bibliographystyle{IEEEbib}
\bibliography{citations}

\vfill\pagebreak

\end{document}